\documentclass[10pt,twocolumn,letterpaper]{article}

\usepackage{iccv}
\usepackage{times}
\usepackage{epsfig}
\usepackage{graphicx}
\usepackage{amsmath}
\usepackage{amssymb}

\usepackage[accsupp]{axessibility}  %
\usepackage{mypackages}
\def\datasetname{TREK-150} 
\def\datasetlink{\href{https://machinelearning.uniud.it/datasets/trek150/}{\footnotesize\texttt{https://machinelearning.uniud.it/datasets/trek150/}}}
\def\videolink{\href{https://youtu.be/oX1nICHgEJM}{\footnotesize\texttt{https://youtu.be/oX1nICHgEJM}}}

\newcommand{\AF}[1]{\textcolor{black}{#1}}
\newcommand{\pgraph}[1]{\paragraph{#1}}

\usepackage[breaklinks=true,bookmarks=false]{hyperref}

\iccvfinalcopy %

\def\iccvPaperID{15} %

\ificcvfinal\fi

\begin{document}

\title{Is First Person Vision Challenging for Object Tracking?}

\author{Matteo Dunnhofer$^{\bullet}$\\
\and
Antonino Furnari$^{\star}$
\and
Giovanni Maria Farinella$^{\star}$
\and
Christian Micheloni$^{\bullet}$ \and
$^{\bullet}$Machine Learning and Perception Lab, University of Udine, Udine, Italy \\
$^{\star}$Image Processing Laboratory, University of Catania, Catania, Italy
}

\maketitle
\ificcvfinal\thispagestyle{empty}\fi

\begin{abstract}
   Understanding 
human-object interactions is 
fundamental 
in First Person Vision (FPV). 
Tracking algorithms \AF{which}
follow the objects \AF{manipulated by}
the camera wearer 
\AF{can provide useful cues to effectively model such interactions.}
Despite 
\AF{a few previous attempts to exploit trackers in FPV applications, a methodical analysis of the performance of state-of-the-art visual trackers in this domain is still missing.}
In this short paper, we provide a recap of the first systematic study of object tracking in FPV. Our work extensively analyses the performance of recent and baseline FPV trackers with respect to different aspects. This is achieved through \datasetname, a novel benchmark dataset composed of 150 densely annotated video sequences.
The results suggest that more research efforts should be devoted to this problem so that tracking could benefit FPV tasks.

\end{abstract}

\section{Introduction}
Understanding the interactions between a camera wearer and the surrounding objects is a fundamental problem in First Person Vision (FPV)~\AF{\cite{EK55,RULSTMpami,damen2016you}}.
\AF{To model such interactions,}
the continuous knowledge of where \AF{an object of interest} is located inside the video frame \AF{is advantageous}.
\begin{figure}[t]%
\centering
\includegraphics[width=\columnwidth]{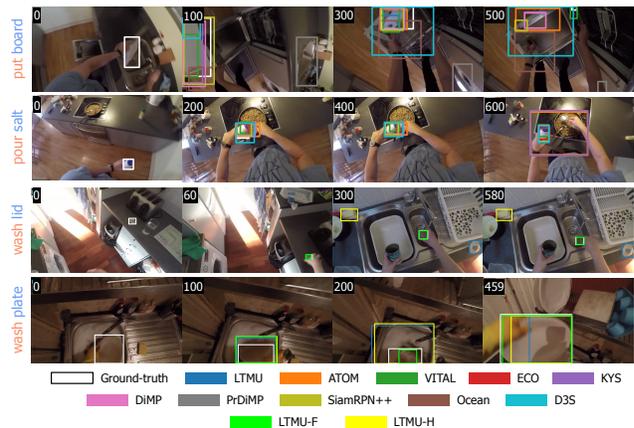}
\caption{Qualitative results of some of the studied trackers on the proposed \datasetname\ dataset. The first 2 rows of images show the qualitative performance of 10 of the selected generic-object trackers, while the last 2 rows show the results of the baseline FPV trackers LTMU-F and LTMU-H in comparison with LTMU. 
For each sequence, the action performed by the camera wearer is also reported (verb in orange, noun in blue). 
For a better visualization, a video can be found at \videolink.}
\label{fig:examples}
\vspace{-1.25em}
\end{figure}
To obtain such an information, most FPV solutions rely on
object detection models that evaluate video frames 
\AF{independently.}
This \AF{paradigm}
has the drawback of ignoring all the temporal information coming from the object appearance and motion contained in consecutive video frames \AF{and generally requires a higher computational cost due to the repeated detection process on every frame.}
In contrast, visual object tracking aims to exploit past information about a target \AF{to}
infer its position and shape in the next frames of a video~\AF{\cite{Maggio2011}}. 
\AF{The benefits of tracking in FPV have been explored by a few previous}
works  %
\AF{to predict future active objects~\cite{Furnari2017}, analyze social interactions~\cite{Aghaei2016icpr}, improve the performance of hand detection for rehabilitation purposes~\cite{Visee2020}, locate hands and capture their movements for action recognition~\cite{kapidis2019egocentric} and human-object interaction forecasting~\cite{liu2020forecasting}.}
Such works have proposed customized approaches to track specific targets like people, people faces, or hands \AF{from the FPV perspective}. 
A solution specifically designed to track arbitrary objects in egocentric videos is still missing.
This is due to the FPV challenges such as camera motion, persistent occlusion, significant scale and state changes, as well as motion blur, which prevent state-of-the-art visual trackers \cite{KCF,SiamFC,DiMP,Dunnhofer2020accv} to track well generic objects.
To support such a claim, we analyzed in-depth the problem of visual object tracking in \AF{the} FPV \AF{domain}. Particularly, we investigated the accuracy and speed performance of both non-FPV and FPV visual trackers, using standard and new performance evaluation strategies. We followed the visual tracking community's standard practice of building an accurate dataset for evaluation \cite{OTB,VOT2019} and proposed a novel visual tracking benchmark, \datasetname\ (TRacking-Epic-Kitchens-150) \cite{TREK150}, which is obtained from the large and challenging FPV dataset EPIC-KITCHENS-55 (EK-55) \cite{EK55}. \datasetname\ provides 150 video sequences densely annotated with the bounding boxes of a %
target object the camera wearer interacts with. Additionally, sequences
\AF{have been}
labelled with attributes that identify the visual changes 
\AF{the object and scene are undergoing, the class of the target object}
and the action the person is performing. 
The results of the study showed that FPV offers challenging tracking scenarios for the most recent \AF{and} accurate trackers and even for FPV trackers. \AF{Considering the potential impact of tracking on FPV, we suggest that more research efforts should be devoted to the considered task, for which we believe the proposed \datasetname\ benchmark will be a key research tool.}
Please refer to \cite{TREK150} for the extended version of this abstract, including more details and results.
Resources regarding \datasetname\ are available at \datasetlink.

\section{The \datasetname\ Benchmark Dataset}

The proposed \datasetname\ dataset \cite{TREK150} is composed of 150 video sequences sampled from the EK-55 dataset \cite{EK55}.
In each video, a single target object is labeled with an axis-aligned bounding box which encloses the visible parts of the object. Dense annotations were generated because it has been shown that they are required for visual tracking benchmarking~\cite{VOT2019}.
To be compliant with other tracking challenges, every sequence is additionally labeled with one or more of 17 attributes describing the visual variability of the target in the sequence. They include 13 standard tracking attributes, plus 4 additional ones (High Resolution, Head Motion, 1-Hand Interaction, 2-Hands Interaction) which are introduced to characterize FPV sequences.
\AF{In addition, 20 action verbs and 34 noun attributes are used to indicate the action performed by the camera wearer and the class of the target.}
Despite being a subset of EK-55, \datasetname\ reflects its long-tail distribution of labels.
Some qualitative examples of the video sequences are shown in Figure \ref{fig:examples}.

\section{Trackers}
We considered 31 trackers to represent
different state-of-the-art approaches to generic object tracking, \AF{for instance with respect to the} %
matching strategy, type of image representations, learning strategy, etc.
\AF{Specifically, in the analysis we} have included short-term trackers based on correlation-filters with both hand-crafted features and deep features. We also considered deep siamese networks, tracking-by-detection methods, as well as trackers based on target segmentation representations, meta-learning, and fusion strategies.
Long-term trackers have been also taken into account in the study. 

In addition to the generic object trackers, we developed 2 baseline FPV trackers that combine the LTMU tracker \cite{LTMU} with (i) the EK-55 trained Faster-R-CNN \cite{EK55} and (ii) the Faster-R-CNN-based hand-object detector \cite{Shan2020}, referred as LTMU-F and LTMU-H respectively.
These baseline trackers exploit the respective detectors as object re-detection modules according to the LTMU scheme \cite{LTMU}. In short, the re-detection happens when a verification module notices that the tracker is not following the correct target. In such a case, the module triggers the execution of the respective FPV detector which proposes candidate locations of the target object. Each of the candidates is evaluated by the verification module, and the location with highest confidence is used to re-initialize the tracker.

\section{Evaluation}
\label{sec:exp}

We used the one-pass evaluation (OPE) protocol \cite{OTB} that consists in initializing a tracker with the target's ground-truth bounding box in the first frame and let the tracker run \AF{on} every other frame until the end of a video.
To obtain a more robust evaluation, we employed the multi-start evaluation (MSE) protocol \cite{VOT2020} which defines different points of initialization along a video. A tracker is initialized with the ground-truth in each point and let run either forward or backward in time (depending on the longest sub-sequence yielded by the initialization point) until the end of the sub-sequence.
Since many FPV tasks require real-time computation, we evaluated the ability of trackers to provide their object localization in such a setting by following the RTE protocol \cite{VOT2017,Li2020}. It runs an algorithm considering its running time and skips all the frames, that occur regularly according to the frame rate, which appeared in between the algorithm's execution start and end times.

\begin{figure*}[t]%
\centering
\includegraphics[width=\linewidth]{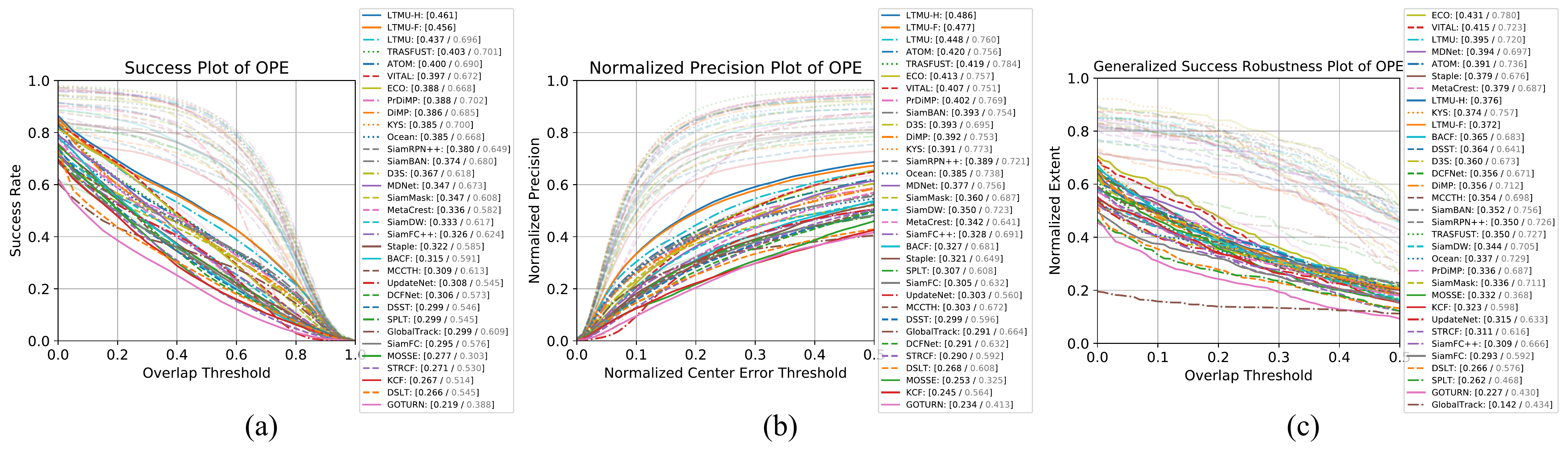}
\caption{Performance of the selected trackers on the proposed \datasetname\ benchmark under the OPE protocol. The curves in solid colors report the performance of the 33 benchmarked trackers on \datasetname,\ whereas the curves overlaid in semi-transparent colors outline the performance obtained by the same trackers on the standard OTB-100~\cite{OTB} dataset. \AF{In brackets, next to the trackers' names, we report the SS, NPS and GR values achieved on \datasetname\ (in black) and on OTB-100 \cite{OTB} (in gray)}.  
As can be noted, all the trackers \AF{exhibit} a significant performance drop when 
\AF{tested on our challenging FPV benchmark}.
LTMU-H and LTMU-F achieve marginally better performance, while we expect significant boosts to be achievable with a careful design of FPV trackers. }
\label{fig:results}
\end{figure*}

\begin{table*}[t]
\fontsize{6}{7}\selectfont
	\centering
	\caption{Performance achieved by 17 of the benchmarked trackers on \datasetname\ using the RTE protocol.}
	\label{tab:realtime}
	\setlength\tabcolsep{.15cm}
	\begin{tabular}{l | c c c c c c c c c c c c c c c c c }
		\toprule
		
		Metric & Ocean & SiamBAN & SiamRPN++ & DiMP & KYS & ATOM & LTMU & D3S & ECO & GlobalTrack & Staple & MOSSE & LTMU-H & MetaCrest & LTMU-F & VITAL & KCF \\

		\midrule
		
		FPS & 21 & 24 & 23 & 16 & 12 & 15 & 8 & 16 & 15 & 8 & 13 & 26 & 4 & 8 & 4 & 4 & 6 \\
		SS & 0.365 & 0.360 & 0.362 & 0.336 & 0.327 & 0.319 & 0.284 & 0.276 & 0.252 & 0.253 & 0.249 &  0.227 & 0.213 & 0.207 & 0.205 & 0.204 & 0.186 \\
		NPS & 0.358 & 0.366 & 0.356 & 0.331 & 0.317 & 0.312 & 0.257 & 0.263 & 0.231 & 0.227 & 0.236 & 0.190 & 0.174 & 0.175 & 0.161 & 0.165 & 0.157 \\
		GSR & 0.294 & 0.313 & 0.293 & 0.224 & 0.237 & 0.179 & 0.169 & 0.182 & 0.173 & 0.139 & 0.169 &  0.141 & 0.161 & 0.165 & 0.162 & 0.158 & 0.177 \\

		\bottomrule		
\end{tabular}
\end{table*}

\begin{figure*}[t]%
\centering
\includegraphics[width=\linewidth]{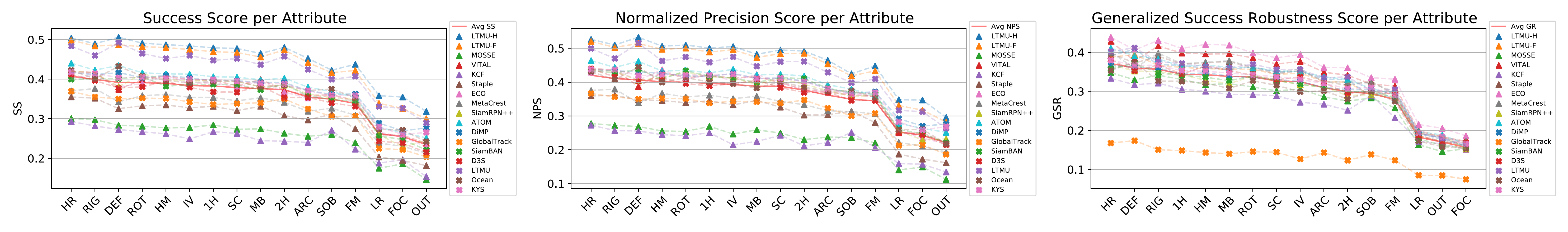}
\caption{SS, NPS, and GSR of 17 of the benchmarked trackers on the sequence attributes of proposed \datasetname\ benchmark under the MSE protocol. The red plain line highlights the average performance. }
\label{fig:resultsattributes}
\end{figure*}

To compare the tracker's predicted bounding boxes with the temporally aligned ground-truth \AF{bounding boxes} we used different measures: the success plot \cite{OTB} which shows the percentage of \AF{predicted} bounding boxes whose intersection-over-union with the ground-truth is larger than a threshold \AF{varied from 0 to 1} (Figure~\ref{fig:results}~(a)); the normalized precision plot \cite{TrackingNet} that \AF{reports}, \AF{for a variety of distance thresholds}, the percentage of bounding boxes whose center points are within a given normalized distance (in pixels) from the ground-truth (Figure \ref{fig:results} (b)); a novel plot which we refer to as generalized success robustness plot (Figure \ref{fig:results} (c)) which measures the normalized extent of a tracking sequence before a failure.
\AF{As summary measures, we report the success score (SS) \cite{OTB}, normalized precision score (NPS) \cite{TrackingNet}, and generalized success robustness (GSR), which are computed as the Area Under the Curve (AUC) of the success plot, the normalized precision plot, and generalized success robustness plot respectively.}

\section{Results}
\label{sec:resuluts}

\begin{table*}[t]
\fontsize{6}{7}\selectfont
	\centering
	\caption{Accuracy results on \datasetname\ of a video-based hand-object detection solution which considers each of the considered trackers as localization method for the object involved in the interaction. As a baseline, we employ the object detection capabilities of the hand-object interaction solution Hands-in-contact \cite{Shan2020}. }
	\label{tab:handsobj}
	\setlength\tabcolsep{.11cm}
	\begin{tabular}{c | c c c c c c c c c c c c c c c c c }
		\toprule
		Hands-in-contact \cite{Shan2020} & LTMU-H & LTMU-F & ATOM & LTMU & Ocean & SiamBAN & SiamRPN++ & MetaCrest & D3S & DiMP & KYS & VITAL & GlobalTrack & MOSSE & ECO & Staple & KCF \\
		\midrule
		0.354 & 0.368 & 0.367 & 0.361 & 0.354 & 0.340 & 0.340 & 0.311 & 0.293 & 0.292 & 0.292 & 0.279 & 0.253 & 0.251 & 0.231 & 0.230 & 0.197 & 0.177 \\

		\bottomrule		
\end{tabular}
\vspace{-0.5em}
\end{table*}

\pgraph{General Remarks.}
Figure \ref{fig:results} \AF{reports} the performance of the selected trackers on \datasetname\ using the OPE protocol, as well their performance on the popular OTB-100 \cite{OTB} benchmark for reference (semi-transparent curves - gray numbers in brackets).
The results are decreased across all measures \AF{when considering the challenging FPV scenario of \datasetname}, demonstrating that the particular characteristics of FPV introduce challenging scenarios for visual trackers.
Some qualitative examples of the trackers' performance are shown in Figure \ref{fig:examples}.
Trackers \AF{based on deep learning} perform better in SS and NPS than those based on hand-crafted features. Among the formers, those that leverage online adaptation mechanisms are more accurate than \AF{the ones based on} single-shot instances.
The generalized success robustness plot in Figure \ref{fig:results}(c) reports a different ranking of the trackers, showing that more spatially accurate trackers are not always able to maintain longer reference to targets.
The proposed FPV trackers LTMU-H and LTMU-F are better in SS and NPS, while they lose some performance in GSR.
Such outcome shows that adapting a state-of-the-art method to FPV allows to marginally improve results. We expect significant performance improvements to be achievable by a tracker accurately designed to tackle the FPV challenges.

Table \ref{tab:realtime} \AF{reports} the FPS performance of the trackers and the SS, NPS, and GSR scores achieved under the RTE protocol. It can be noted the performance of all trackers is decreased when considering their running time in relation to the frame rate. These results suggest that efforts should be made to make trackers accurate in real-time applications.

\pgraph{Conditions That Influence The Trackers.}
Figure \ref{fig:resultsattributes} \AF{reports} the SS, NPS, and GSR scores of trackers, computed with the MSE protocol, with respect to the 
attributes used to characterize the visual variability of the sequences.
Full occlusion~(FOC), out of view~(OUT) and the small size of targets~(LR) \AF{are} the most difficult situations. The fast motion of targets~(FM) and the presence of similar objects~(SOB) are also critical factors that cause drops in performance. 
Trackers are less vulnerable to rotations~(ROT) and to the illumination variation~(IV). 
Tracking objects held with two hands~(2H) is more difficult than tracking objects held with a single hand~(1H),
while trackers are quite robust to \AF{head motion}~(HM) and objects appearing in larger size~(HR).
Considering the performance with respect to actions, it results that those \AF{mainly} causing a spatial \AF{displacement} 
of the target (e.g. ``move'', ``store'', ``check'') have less impact.
Actions that change the state, shape, or aspect ratio of an object (e.g. ``remove'', ``squeeze'', ``cut'', ``attach'') generate harder tracking scenarios.
\AF{Also the sequences including the ``wash'' verb lead to poor performance.}
With respect to the associated noun labels, we have that 
\AF{rigid, regular-sized objects such as ``pan'', ``kettle'', ``bowl'', ``plate'', and ``bottle'' are among the ones associated with high average scores.}
Other rigid objects such as ``knife'', ``spoon'', ``fork'' and ``can'' are harder
to track instead, probably due to their particularly thin shape and the light reflectance they are \AF{easily} subject to.  
\AF{Deformable objects such as ``sponge'', ``onion'',  ``cloth'' and ``rubbish'' are also difficult to track.}

\pgraph{Impact of Trackers in FPV.}
Despite we demonstrated that FPV is challenging for current trackers, \AF{we assessed whether these already offer an advantage in the FPV domain to obtain information about the objects' locations and movements in the scene.
To this aim, we evaluated the performance of a Faster R-CNN instance trained on EK-55 \cite{EK55} when used as a naive tracking baseline.} Such a solution achieved an SS, NPS, and GSR of 0.323, 0.369, 0.044, by running at 1 FPS. Comparing these results with the ones presented in Figure \ref{fig:results}, we clearly notice that trackers, if properly initialized by a detection module, can deliver faster, more accurate and much more temporally long object localization than detectors.
We additionally evaluated the accuracy of a video-based hand-object interaction detection solution \cite{Shan2020} whose object localisation is given by a tracker rather than a detector. The tracker is initialized with the object detector's predicted bounding box at the first detection of the hand-object interaction, and let run until its end. By this setting, we created a ranking of the trackers which is presented in Table \ref{tab:handsobj}. The results demonstrate that stronger trackers can improve the accuracy and efficiency of current detection-based methodologies \cite{Shan2020}. Interestingly, the trackers' ranking differs from what shown in Figure \ref{fig:results}, suggesting that trackers can manifest other capabilities when deployed into application scenarios.

\AF{Given these preliminary results, we hence expect that trackers will likely gain more importance in FPV as new methodologies explicitly considering the first person point of view are investigated.}

{\fontsize{5.25}{6.25}\selectfont
\bibliographystyle{ieee_fullname}
\bibliography{egbib}
}

\end{document}